\documentclass[11pt,a4paper]{article}
\usepackage[hyperref]{emnlp2018}
\usepackage{times}
\usepackage{latexsym}
\usepackage{multirow}
\usepackage{url}
\usepackage{comment}
\usepackage{amsmath}

\usepackage{xcolor}

\usepackage{graphicx}

\aclfinalcopy

\title{Improving Cross-Lingual Word Embeddings by Meeting in the Middle}

\author{Yerai Doval \\
  Grupo COLE, Escola Superior \\ de Enxe\~{n}ar\'ia Inform\'atica \\
  Universidade de Vigo, Spain \\
  {\tt yerai.doval@uvigo.es} \\\And
  Jose Camacho-Collados, Luis Espinosa-Anke \\ \textbf{and Steven Schockaert} \\
  School of Computer Science 
and Informatics \\
  Cardiff University, UK \\
  {\tt camachocolladosj@cardiff.ac.uk} \\
  {\tt espinosa-ankel@cardiff.ac.uk}  \\
  {\tt schockaerts1@cardiff.ac.uk}  \\ }

\date{}

\begin{document}
\maketitle
\begin{abstract}
  
  Cross-lingual word embeddings are becoming increasingly important in multilingual NLP. 
  Recently, it has been shown that 
  these embeddings can be effectively learned by aligning two disjoint monolingual vector spaces through linear transformations
  , using no more than a small bilingual dictionary as supervision.
  In this work, we propose to apply an additional transformation after the initial alignment step, which moves cross-lingual synonyms towards a middle point between them. By applying this transformation our aim is to obtain a better cross-lingual integration of the vector spaces. In addition, and perhaps surprisingly, the monolingual spaces also improve by this transformation. 
  This is in contrast to the original alignment, which is typically learned such that the structure of the monolingual spaces is preserved.
  Our experiments confirm that the resulting cross-lingual embeddings outperform 
  state-of-the-art models 
  in both monolingual and cross-lingual evaluation tasks. 
\end{abstract}

\section{Introduction}

Word embeddings are one of the most widely used resources in NLP, as they have proven to be of enormous importance for modeling linguistic phenomena in both supervised and unsupervised settings.
In particular, the representation of words in cross-lingual vector spaces (henceforth, \textit{cross-lingual word embeddings}) is quickly gaining in popularity. 
One of the main reasons is that they play a crucial role in transferring knowledge from one language to another, specifically in downstream tasks such as information retrieval \cite{vulic2015monolingual}, entity linking \cite{tsai2016cross} and text classification \cite{mogadala2016bilingual}, while at the same time providing improvements in multilingual NLP problems such as machine translation~\cite{zou2013bilingual}.

There exist different approaches for obtaining these cross-lingual embeddings. One of the most successful methodological directions, which constitutes the main focus of this paper, attempts to learn bilingual embeddings via a two-step process: first, word embeddings are trained on monolingual corpora and then the resulting monolingual spaces are aligned by taking advantage of bilingual dictionaries~\cite{mikolov2013exploiting,faruqui2014improving,xing2015normalized}. 

These \textit{alignments} are generally modeled as linear transformations, which are constrained such that the structure of the initial monolingual spaces is left unchanged. This can be achieved by imposing an orthogonality constraint on the linear transformation \cite{xing2015normalized,artetxe2016learning}. 
Our hypothesis in this paper is that such approaches can be further improved, as they rely on the assumption that the internal structure of the two monolingual spaces is identical. In reality, however, this structure is influenced by language-specific phenomena, e.g., the fact that Spanish distinguishes between masculine and feminine nouns \cite{davis2015does} as well as the specific biases of the different corpora from which the monolingual spaces were learned. Because of this, monolingual embedding spaces are not isomorphic \cite{sogaard2018limitations,yuva2018generalizing}. On the other hand, simply dropping the orthogonality constraints leads to overfitting, and is thus not effective in practice.

The solution we propose is to start with existing state-of-the-art alignment models \cite{artetxe-labaka-agirre:2017:Long,conneau2018word}, and to apply a further transformation to the resulting initial alignment. For each word $w$ with translation $w'$, this additional transformation aims to map the vector representations of both $w$ and $w'$ onto their average, thereby creating a cross-lingual vector space which intuitively corresponds to the average of the two aligned monolingual vector spaces. Similar to the initial alignment, this mapping is learned from a small bilingual lexicon. 

Our experimental results show that the proposed additional transformation does not only benefit cross-lingual evaluation tasks, but, perhaps surprisingly, also monolingual ones. In particular, we perform an extensive set of experiments on standard benchmarks for bilingual dictionary induction and monolingual and cross-lingual word similarity, as well as on an extrinsic task: cross-lingual hypernym discovery. 

Code and pre-trained embeddings to reproduce our experiments and to apply our model to any given cross-lingual embeddings are available at \url{https://github.com/yeraidm/meemi}. 

\section{Related Work}

Bilingual word embeddings have been extensively studied in the literature in recent years. Their nature varies with respect to the supervision signals used for training \cite{upadhyay2016cross,ruder2017survey}. Some common signals to learn bilingual embeddings come from parallel \cite{hermann2014multilingual,luong2015bilingual,levy2017strong} or comparable corpora \cite{vulic2015bilingual,sogaard2015inverted,vulic2016bilingual}, or lexical resources such as WordNet, ConceptNet or BabelNet \cite{conceptnet2017,Mrksik:17tacl, goikoetxea2018bilingual}. However, these sources of supervision may be scarce, limited to certain domains or may not be directly available for certain language pairs. 

Another branch of research exploits pre-trained monolingual embeddings with weak signals such as bilingual lexicons for learning bilingual embeddings \cite{mikolov2013exploiting,faruqui2014improving,ammar2016massively,artetxe2016learning}. \newcite{mikolov2013exploiting} was one of the first attempts into this line of research, applying a linear transformation in order to map the embeddings from one monolingual space into another. They also noted that more sophisticated approaches, such as using multilayer perceptrons, do not improve with respect to their linear counterparts. \newcite{xing2015normalized} built upon this work by normalizing word embeddings during training and adding an orthogonality constraint. In a complementary direction, \newcite{faruqui2014improving} put forward a technique based on canonical correlation analysis to obtain linear mappings for both monolingual embedding spaces into a new shared space. \newcite{artetxe2016learning} proposed a similar linear mapping to \newcite{mikolov2013exploiting}, generalizing it and providing theoretical justifications which also served to reinterpret the methods of \newcite{faruqui2014improving} and \newcite{xing2015normalized}.
\newcite{smith2017offline} further showed how orthogonality was required to improve the consistency of bilingual mappings, making them more robust to noise.
Finally, a more complete generalization providing further insights on the linear transformations used in all these models can be found in \newcite{artetxe2018generalizing}. 

These approaches generally require large bilingual lexicons to effectively learn multilingual embeddings \cite{artetxe-labaka-agirre:2017:Long}. Recently, however, alternatives which only need very small dictionaries, or even none at all, have been proposed to learn high-quality embeddings via linear mappings \cite{artetxe-labaka-agirre:2017:Long,conneau2018word}. More details on the specifics of these two approaches can be found in Section \ref{method:linearmappings}. These models 
have in turn paved the way for the development of machine translation systems which do not require any parallel corpora  \cite{artetxe2018iclr,lample2018unsupervised}.  Moreover, the fact that such approaches only need monolingual embeddings, instead of parallel or comparable corpora, makes them easily adaptable to different domains (e.g., social media or web corpora). 

In this paper we build upon these state-of-the-art approaches by applying an additional transformation, which aims to map each word and its translation onto the average of their vector representations. This strategy bears some resemblance with the idea of learning meta-embeddings \cite{DBLP:conf/acl/YinS16}. Meta-embeddings are vector space representations which aggregate several pre-trained word embeddings from a given language (e.g., trained using different corpora and/or different word embedding models). Empirically it was found that such meta-embeddings can often outperform the individual word embeddings from which they were obtained. In particular, it was recently argued that word vector averaging can be a highly effective approach for learning such meta-embeddings \cite{coates2018frustratingly}. The main difference between such approaches and our work is that because we rely on a small dictionary, we cannot simply average word vectors, since for most words we do not know the corresponding translation. Instead, we train a regression model to predict this average word vector from the vector representation of the given word only, i.e., without using the vector representation of its translation.

\section{Methodology}

Our approach for improving cross-lingual embeddings consists of three main steps, where the first two steps are the same as in existing methods. In particular,  given two monolingual corpora, a word vector space is first learned independently for each language. This can be achieved with common word embedding models, e.g., Word2vec \cite{Mikolovetal:2013}, GloVe \cite{pennington2014glove} or FastText \cite{bojanowski2017enriching}. Second, a linear alignment strategy is used to map the monolingual embeddings to a common bilingual vector space (Section \ref{method:linearmappings}). Third, a final transformation is applied on the aligned embeddings so the word vectors from both languages are refined and further integrated with each other (Section \ref{neural}). 
This third step is the main contribution of our paper.

\subsection{Aligning monolingual spaces}
\label{method:linearmappings}

Once the monolingual word embeddings have been obtained, a linear transformation is applied in order to integrate them into the same vector space. This linear transformation is generally carried out using a supervision signal, typically in the form of a bilingual dictionary. In the following we explain two state-of-the-art models performing this linear transformation.

\paragraph{\texttt{VecMap}  \cite{artetxe-labaka-agirre:2017:Long}.} 
\texttt{VecMap} uses an orthogonal transformation over normalized word embeddings. 
An iterative two-step procedure is also implemented in order to avoid the need of starting with a large seed dictionary (e.g., in the original paper it was tested with a very small bilingual dictionary of just 25 pairs). 
In this procedure, first, the linear mapping is estimated using a small bilingual dictionary, and then, this dictionary is augmented by applying the learned transformation to new words from the source language. Lastly, the process is repeated until some convergence criterion is met. 

\paragraph{\texttt{MUSE} \cite{conneau2018word}.} 
In this case, the transformation matrix is learned through an iterative Procrustes alignment~\cite{schonemann1966generalized}.\footnote{Very recently, \newcite{yuva2018generalizing} showed that projecting both monolingual embedding spaces onto a third space (instead of directly onto each other) using a generalized Procrustes analysis facilitates the learning of alignments.
} The anchor points needed for this alignment can be obtained either through a supplied bilingual dictionary or through an unsupervised model. This unsupervised model is trained using adversarial learning to obtain an initial alignment of the two monolingual spaces, which is then refined by the Procrustes alignment using the most frequent words as anchor points.
A new distance metric for the embedding space, referred to as cross-domain similarity local scaling, is also introduced. This metric, which takes into account the nearest neighbors of both source and target words, was shown to better handle high-density regions of the space, thus alleviating the hubness problem of 
word embedding models~\cite{radovanovic2010hubs,dinu2015improving}, which arises when a few points (known as hubs) become the nearest neighbors of many other points in the embedding space.

\subsection{Meeting in the middle}
\label{neural}

After 
the initial alignment of the monolingual word embeddings,
our proposed method leverages an additional linear model to refine the resulting bilingual word embeddings. 
This is because the methods presented in the previous section apply constraints to ensure that the structure of the monolingual embeddings is largely preserved. As already mentioned in the introduction, conceptually this may not be optimal, as embeddings for different languages and trained from different corpora can be expected to be structured somewhat differently. Empirically, as we will see in the evaluation, after applying methods such as  \texttt{VecMap} and \texttt{MUSE} there still tend to be significant gaps between the vector representations of words and their translations. Our method directly attempts to reduce these gaps by moving each word vector towards the middle point between its current representation and the representation of its translation. In this way, by bringing the two monolingual fragments of the space closer to each other, we can expect to see an improved performance on cross-lingual evaluation tasks such as bilingual dictionary induction. Importantly, the internal structure of the two monolingual fragments themselves is also affected by this step. By averaging between the representations obtained from different languages, we hypothesize that the impact of language-specific phenomena and corpus specific biases will be reduced, thereby ending up with more ``neutral'' monolingual embeddings.

In the following, we detail our methodological approach.
First, we leverage the same bilingual dictionary that was used to obtain the initial alignment (Section \ref{method:linearmappings}). Specifically, let $D=\{(w,w')\}$ be the given bilingual dictionary, where $w \in V$ and $w' \in V'$, with $V$ and $V'$ representing the vocabulary of the first and second language, respectively. For pairs $(w,w') \in D$, we can simply compute the corresponding average vector $\vec{\mu}_{w,w'}=\frac{\vec{v}_w+\vec{v}_{w'}}{2}$. Then, using the pairs in $D$ as training data, we learn a linear mapping $X$ such that $X \vec{v}_w \approx \vec{\mu}_{w,w'}$ for all $(w,w')\in D$. This mapping $X$ can then be used to predict the averages for words outside the given dictionary. 
To find the mapping $X$, we solve the following least squares linear regression problem: 
\begin{equation}\label{eqRegressionFormula}
    E=\sum_{(w,w') \in D} \|X\vec{w}-\vec{\mu}_ {w,w'}\|^2
\end{equation}

\noindent Similarly, for the other language, we separately learn a mapping $X'$ such that $X' \vec{v}_{w'} \approx \vec{\mu}_{w,w'}$. 

It is worth pointing out that we experimented with several variants of this linear regression formulation. For example, we also tried using a multilayer perceptron to learn non-linear mappings, and we experimented with several regularization terms to penalize mappings that deviate too much from the identity mapping. None of these variants, however, were found to improve on the much simpler formulation in \eqref{eqRegressionFormula}, which can be solved exactly and efficiently.
Furthermore, one may wonder whether the initial alignment is actually needed, since e.g., \newcite{coates2018frustratingly} obtained high-quality meta-embeddings without such an alignment set. However, when applying our approach directly to the initial monolingual non-aligned embedding spaces, we obtained results which were competitive but slightly below the two considered alignment strategies.

\section{Evaluation}

We test our bilingual embedding refinement approach on both intrinsic and extrinsic tasks.  
In Section \ref{training} we describe the common training setup for all experiments and language pairs. 
The languages we considered 
are English, Spanish, Italian, German and Finnish. 
Throughout all the experiments we use publicly available resources in order to make comparisons and reproducibility of our experiments easier.

\subsection{Cross-lingual embeddings training}
\label{training}

\noindent \textbf{Corpora.} In our experiments we make use of web-extracted corpora. 
For English we use the 3B-word UMBC WebBase Corpus~\cite{han2013umbc}, 
while we chose the Spanish Billion Words Corpus~\cite{cardellinoSBWCE} for Spanish. 
For Italian and German, we use the itWaC and sdeWaC corpora from the WaCky project~\cite{baroni2009wacky}, containing 2 and 0.8 billion words, respectively.\footnote{UMBC, Spanish Billion-Words and ItWaC are the official corpora of the hypernym discovery SemEval task (Section \ref{hypernym_discovery}) for English, Spanish and Italian, respectively. } 
Lastly, for Finnish, we use the Common Crawl monolingual corpus from the Machine Translation of News Shared Task 2016\footnote{\url{http://www.statmt.org/wmt16/translation-task.html}}, composed of 2.8B words. All corpora are tokenized and lowercased.

\smallskip
\noindent \textbf{Monolingual embeddings.} The monolingual word embeddings are trained with the Skipgram model from FastText \cite{bojanowski2017enriching} on the corpora described above. The dimensionality of the vectors was set to 300, with the default FastText hyperparameters. 

\begin{table*}[!t]
\centering
\tiny
\renewcommand{\arraystretch}{1.15}
\setlength{\tabcolsep}{2.25pt}
\resizebox{\textwidth}{!}{
\begin{tabular}{l|rrr|rrr|rrr|rrr}
\multicolumn{1}{c|}{\multirow{2}{*}{Model}} & \multicolumn{3}{c|}{\textbf{\texttt{EN-ES}}}                                                            & \multicolumn{3}{c|}{\textbf{\texttt{EN-IT}}}                                                            & \multicolumn{3}{c|}{\textbf{\texttt{EN-DE}}}                                                            & \multicolumn{3}{c}{\textbf{\texttt{EN-FI}}}                                                           \\ \cline{2-13} 
\multicolumn{1}{c|}{}                       & \multicolumn{1}{c}{$P@1$} & \multicolumn{1}{c}{$P@5$} & \multicolumn{1}{c|}{$P@10$} & \multicolumn{1}{c}{$P@1$} & \multicolumn{1}{c}{$P@5$} & \multicolumn{1}{c|}{$P@10$} & \multicolumn{1}{c}{$P@1$} & \multicolumn{1}{c}{$P@5$} & \multicolumn{1}{c|}{$P@10$} & \multicolumn{1}{c}{$P@1$} & \multicolumn{1}{c}{$P@5$} & \multicolumn{1}{c}{$P@10$} \\ \hline
\textbf{\texttt{VecMap}}                                      & 36.0                      & 59.8                       & 65.6                        & 35.5                       & 57.2                       & 63.9                        & 31.7                       & 54.2                       & 60.2                        & 17.2                       & 36.4                       & 43.7                       \\
\textbf{\texttt{VecMap}}$_\mu$                                   & \textbf{37.8}                      & \textbf{61.5}                       & \textbf{67.1}                        & \textbf{36.3}                       & \textbf{59.2}                       & \textbf{66.3}                        & \textbf{33.5}                       & \textbf{57.3}                       & \textbf{61.7}                        & \textbf{18.5}                       & \textbf{40.9}                       & \textbf{48.3}                       \\ \hline
\textbf{\texttt{MUSE}}                                        & 37.1                      & 59.0                       & 65.2                        & 36.3                       & 57.3                       & 62.9                        & 32.5                       & 53.7                       & 59.0                        & 18.2                       & 35.2                       & 42.4                       \\
\textbf{\texttt{MUSE}}$_\mu$                                     & \textbf{38.3}                      & \textbf{62.3}                       & \textbf{67.2}                        & \textbf{37.0}                       & \textbf{59.0}                       & \textbf{65.7}                        & \textbf{33.7}                       & \textbf{57.0}                       & \textbf{62.2}                        & \textbf{19.4}                       & \textbf{41.1}                       & \textbf{49.0}                      
\end{tabular}
}
\caption{Bilingual dictionary induction results. Precision at $k$ ($P@K$) performance for Spanish (ES), Italian (IT), German (DE) and Finnish (FI), using English (EN) as source language. 
}
\label{tab:dictinduction}
\end{table*}

\smallskip
\noindent \textbf{Bilingual dictionaries.} 
We use the bilingual dictionaries packaged together by \newcite{artetxe-labaka-agirre:2017:Long}, each one conformed by 5000 word translations.
They are used both for the initial bilingual mappings and then again for our linear transformation. 

\smallskip
\noindent\textbf{Initial mapping.}  Following previous works, for the purpose of obtaining the initial alignment, English is considered as source language and the remaining languages are used as target. 
We make use of the open-source implementations of \textbf{\texttt{VecMap}}\footnote{\url{github.com/artetxem/vecmap}} \cite{artetxe-labaka-agirre:2017:Long} and \textbf{\texttt{MUSE}}\footnote{\url{github.com/facebookresearch/MUSE}} \cite{conneau2018word}, which constitute strong baselines for our experiments (cf.\ Section \ref{method:linearmappings}). Both of them were used with the recommended parameters and in their supervised setting, using the aforementioned bilingual dictionaries. 

\smallskip
\noindent\textbf{Meeting in the Middle.} Then, once the initial cross-lingual embeddings are trained, and as explained in Section \ref{neural}, we obtain our linear transformation by using the exact solution to the least squares linear regression problem. 
To this end, we use the same bilingual dictionaries as in the previous step. 
Henceforth, we will refer to our transformed models as \textbf{\texttt{VecMap}$_\mu$} and \textbf{\texttt{MUSE}$_\mu$}, depending on the initial mapping.

\subsection{Experiments}
\label{experiments}

We test our cross-lingual word embeddings in two intrinsic tasks, i.e., bilingual dictionary induction (Section \ref{dictionary_induction}) and word similarity (Section \ref{similarity}), and an extrinsic task, i.e., cross-lingual hypernym discovery (Section \ref{hypernym_discovery}).

\subsubsection{Bilingual dictionary induction}
\label{dictionary_induction}

The dictionary induction task consists in automatically generating a bilingual dictionary from a source to a target language, using as input a list of words in the source language. 

\paragraph{Experimental setting}

For this task, and following previous works, we use the English-Italian test set released by \newcite{dinu2015improving} and those released by \newcite{artetxe-labaka-agirre:2017:Long} for the remaining language pairs. These test sets have no overlap with respect to the training and development sets, and contain around 1900 entries each. 
Given an input word from the source language, word translations are retrieved through a nearest-neighbor search of words in the target language, using cosine distance. 
Note that this gives us a ranked list of candidates for each word from the source language. Accordingly, the performance of the embeddings is evaluated with the precision at $k$ ($P@k$) metric, which evaluates for what percentage of test pairs, the correct answer is among the $k$ highest ranked candidates.

\paragraph{Results}

As can be seen in Table \ref{tab:dictinduction}, our refinement method consistently improves over the baselines (i.e., \texttt{VecMap} and \texttt{MUSE}) on all language pairs and metrics.
The higher scores indicate that the two monolingual embedding spaces become more tightly integrated because of our additional transformation. 
It is worth highlighting here the case of English-Finnish, where the gains obtained in $P@5$ and $P@10$ are considerable.
This might indicate that our approach is especially useful for morphologically richer languages such as Finnish, where the limitations of the previous bilingual mappings are most apparent.

\paragraph{Analysis}

When analyzing the source of errors in $P@1$, we came to similar conclusions as \newcite{artetxe-labaka-agirre:2017:Long}.\footnote{The results on this task are lower than those reported in~\newcite{artetxe-labaka-agirre:2017:Long}. This is due to the different corpora and embedding algorithms used to train the monolingual embeddings. In particular, they use corpora including Wikipedia, which is comparable across languages.} 
Several source words are translated to words that are closely related to the one in the gold reference in the target language; e.g., for the English word \textit{essentially} we obtain \textit{b\'asicamente} (\textit{basically}) instead of \textit{fundamentalmente} (\textit{fundamentally}) in Spanish, both of them closely related, or the closest neighbor for \textit{dirt} being \textit{mugre} (\textit{dirt}) instead of \textit{suciedad} (\textit{dirt}), which in fact was among the five closest neighbors. We can also find multiple examples of the higher performance of our models compared to the baselines. For instance, in the English-Spanish cross-lingual models, after the initial alignment, we can find that \textit{seconds} has \textit{minutos} (\textit{minutes}) as nearest neighbour, but after applying our additional transformation, \textit{seconds} becomes closest to \textit{segundos} (\textit{seconds}). Similarly, \textit{paint} initially has \textit{tintado} (\textit{tinted}) as the closest Spanish word, and then \textit{pintura} (\textit{paint}). 

\begin{table*}[!t]
\centering
\Large
\renewcommand{\arraystretch}{1.4}
\setlength{\tabcolsep}{3.0pt}
\resizebox{\textwidth}{!}{
\begin{tabular}{l|rr|rr|rr|rr||rr|rr||rr|rr||rr|rr|rr}
\multicolumn{1}{c|}{\multirow{3}{*}{Model}} & \multicolumn{8}{c||}{\textbf{\texttt{English}}}                                                                                                                                                                                                      & \multicolumn{4}{c||}{\texttt{\textbf{Spanish}}}                                                                                    & \multicolumn{4}{c||}{\texttt{\textbf{Italian}}}                                                                                    & \multicolumn{6}{c}{\texttt{\textbf{German}}}                                                                                                                                             \\ \cline{2-23} 
\multicolumn{1}{c|}{}                       & \multicolumn{2}{c|}{{\bf SemEval}}                          & \multicolumn{2}{c|}{{\bf WordSim}}                           & \multicolumn{2}{c|}{{\bf SimLex}}                         & \multicolumn{2}{c||}{{\bf RG-65}}                                & \multicolumn{2}{c|}{{\bf SemEval}}                          & \multicolumn{2}{c||}{{\bf RG-65}}                               & \multicolumn{2}{c|}{{\bf SemEval}}                         & \multicolumn{2}{c||}{{\bf WordSim}}                           & \multicolumn{2}{c|}{{\bf SemEval}}                         & \multicolumn{2}{c|}{{\bf WordSim}}                           & \multicolumn{2}{c}{{\bf RG-65}}                               \\ \cline{2-23} 
\multicolumn{1}{c|}{}                       & \multicolumn{1}{c}{$r$} & \multicolumn{1}{c|}{$\rho$} & \multicolumn{1}{c}{$r$} & \multicolumn{1}{c|}{$\rho$} & \multicolumn{1}{c}{$r$} & \multicolumn{1}{c|}{$\rho$} & \multicolumn{1}{c}{$r$} & \multicolumn{1}{c||}{$\rho$} & \multicolumn{1}{c}{$r$} & \multicolumn{1}{c|}{$\rho$} & \multicolumn{1}{c}{$r$} & \multicolumn{1}{c||}{$\rho$} & \multicolumn{1}{c}{$r$} & \multicolumn{1}{c|}{$\rho$} & \multicolumn{1}{c}{$r$} & \multicolumn{1}{c||}{$\rho$} & \multicolumn{1}{c}{$r$} & \multicolumn{1}{c|}{$\rho$} & \multicolumn{1}{c}{$r$} & \multicolumn{1}{c|}{$\rho$} & \multicolumn{1}{c}{$r$} & \multicolumn{1}{c}{$\rho$} \\ \hline
\textbf{\texttt{VecMap}}                                      & 74.1                     & 73.9                        & 67.9                     & 67.0                        & 42.0                     & 40.7                        & 77.8                     &  \bf 77.5                        & 70.0                     & 71.4                        & 86.6                    & 88.0                        & 67.2                     &  \bf 69.0                        & 64.0                     & 66.9                        & 70.1                     & 70.1                        & 72.7                     & 72.2                        & 80.2                     & 79.7                       \\

\textbf{\texttt{VecMap}}$_\mu$  &	 \bf 75.0	&  \bf 74.8 & 	 \bf 70.5 & 	 \bf 70.1 & 	 \bf 43.8 & 	 \bf 41.8 & 	 \bf 78.0 & 	76.6 & 	 \bf 71.5 &  \bf 	72.1 & 	 \bf 87.6 & 	 \bf 89.4 & 	 \bf 68.4 & 	68.9 & 	 \bf 65.3 & 	 \bf 67.3 & 	 \bf 70.9 & 	 \bf 70.7 & 	72.7 &  \bf	72.4 & 	 \bf 81.0 &  \bf 	81.3 \\

\hline

\textbf{\texttt{MUSE}}                                        & 74.2                     & 74.2                        & 68.3                     & 67.6                        & 42.6                     & 41.5                        &  \bf 78.6                     &  \bf 78.4                        & 70.5                     & 71.9                        & 86.6                    & 88.3                        & 67.4                     &  \bf 69.2                        & 64.1                     & 66.9                        & 69.8                     & 69.8                        &  \bf 72.5                     &  \bf 72.5                        & 80.3                     &  \bf 80.1                       \\

\textbf{\texttt{MUSE}}$_\mu$ &	  \bf 75.0 &	 \bf 	74.8 &	 \bf 	70.8 &	 \bf 	70.4 &	 \bf 	44.2 &	 \bf 	42.4 &		78.3 &		77.5 &	 \bf 	71.8 &	 \bf 	72.3 &	 \bf 	87.7 &	 \bf 	89.3 &	 \bf 	68.6 &		69.1 &	 \bf 	65.3 &	 \bf 	67.2 &	 \bf 	70.4 &	 \bf 	70.2 &		72.2 &		72.1 &		80.3 &		80.0 \\

\end{tabular}}
\caption{Monolingual word similarity results. Pearson ($r$) and Spearman ($\rho$) correlation.}
\label{tab:monosim}
\end{table*}

\subsubsection{Word similarity}
\label{similarity}

We perform experiments on both monolingual and cross-lingual word similarity. In monolingual similarity, models are tested in their ability to determine the similarity between two words in the same language, whereas in cross-lingual similarity the words belong to different languages. While in the monolingual setting the main objective is to test the quality of the monolingual subsets of the bilingual vector space, the cross-lingual setting constitutes a straightforward benchmark to test the quality of bilingual embeddings. 

\paragraph{Experimental setting}

For monolingual word similarity we use the English SimLex-999 \cite{hill2015simlex}, and the language-specific versions of SemEval-17\footnote{The original datasets of SemEval-17 contained also multiwords, but for consistency we use the version containing single words only.}~\cite{semeval2017similarity}, WordSim-353\footnote{WordSim datasets consist of the similarity re-scoring for several languages of \newcite{leviant2015separated}, downloaded from \url{http://leviants.com/ira.leviant/MultilingualVSMdata.html}} \cite{Levetal:2002}, and RG-65 \cite{RG65:1965}. The corresponding cross-lingual datasets from SemEval-18, WordSim-353 and RG-65 were considered for the cross-lingual word similarity evaluation\footnote{The WordSim-353 and RG-65 cross-lingual datasets \cite{camacho2015framework} were downloaded at \url{http://lcl.uniroma1.it/similarity-datasets/}}. Cosine similarity is again used as comparison measure.

\paragraph{Results}

Tables \ref{tab:monosim} and \ref{tab:cross-sim} show the monolingual\footnote{The English results correspond to the averaged performance of the English fragments of English-Spanish, English-Italian and English-German cross-lingual embeddings.} and cross-lingual word similarity results\footnote{The results of the original \texttt{VecMap} in cross-lingual similarity are comparable or better to those reported in \newcite{artetxe-labaka-agirre:2017:Long} on the three datasets used in their evaluation.}, respectively. For both the monolingual and cross-lingual settings, we can notice that our models generally outperform the corresponding baselines. Moreover, in cases where no improvement is obtained, the differences tend to be minimal, with the exception of RG-65, but this is a very small test set for which larger variations can thus be expected. In contrast, there are a few cases where substantial gains were obtained by using our model. This is most notable for English WordSim and SimLex in the monolingual setting.

\begin{table*}[!h]
\centering
\scriptsize
\renewcommand{\arraystretch}{1.2}
\setlength{\tabcolsep}{2.2pt}
\resizebox{0.9\textwidth}{!}{
\begin{tabular}{l|rr|rr||rr|rr||rr|rr|rr}
\multicolumn{1}{c|}{\multirow{3}{*}{Model}} & \multicolumn{4}{c||}{\textbf{\texttt{English-Spanish}}}                                                                                      & \multicolumn{4}{c||}{\textbf{\texttt{English-Italian}}}                                                                                      & \multicolumn{6}{c}{\textbf{\texttt{English-German}}}                                                                                                                                               \\ \cline{2-15} 
\multicolumn{1}{c|}{}                       & \multicolumn{2}{c|}{{\bf SemEval}}                        & \multicolumn{2}{c||}{{\bf RG-65}}                                & \multicolumn{2}{c|}{{\bf SemEval}}                         & \multicolumn{2}{c||}{{\bf WordSim}}                           & \multicolumn{2}{c|}{{\bf SemEval}}                         & \multicolumn{2}{c|}{{\bf WordSim}}                           & \multicolumn{2}{c}{{\bf RG-65}}                               \\ \cline{2-15} 
\multicolumn{1}{c|}{}                       & \multicolumn{1}{c}{$r$} & \multicolumn{1}{c|}{$\rho$} & \multicolumn{1}{c}{$r$} & \multicolumn{1}{c||}{$\rho$} & \multicolumn{1}{c}{$r$} & \multicolumn{1}{c|}{$\rho$} & \multicolumn{1}{c}{$r$} & \multicolumn{1}{c||}{$\rho$} & \multicolumn{1}{c}{$r$} & \multicolumn{1}{c|}{$\rho$} & \multicolumn{1}{c}{$r$} & \multicolumn{1}{c|}{$\rho$} & \multicolumn{1}{c}{$r$} & \multicolumn{1}{c}{$\rho$} \\ \hline
\textbf{\texttt{VecMap}}                                      & 71.7                     &  \bf 71.6                        & 82.1                     & 82.4                        & 69.6                     & 69.6                        & 60.2                     &  \bf 63.1                        & 71.6                     & \multicolumn{1}{r|}{71.3}   & 64.1                     & \multicolumn{1}{r|}{ \bf 65.9}   & 78.1                     & 78.8                       \\

\textbf{\texttt{VecMap}}$_\mu$ & 	71.7 & 	71.3 & 	82.1 & 	 \bf 82.8 & 	 \bf 70.2 & 	 \bf 69.9 & 	 \bf 61.3 & 	63.0 & 	 \bf 72.0 & 	 \bf 71.5 & 	 \bf 64.2 & 	65.4 & 	 \bf 78.6 & 	 \bf 79.7 \\
\hline

\textbf{\texttt{MUSE}}                                        & 72.0                     &  \bf 72.0                        & 81.9                     & 82.3                        & 69.4                     & 69.4                        & 59.9                     & 62.7                        & 70.4                     & \multicolumn{1}{r|}{70.1}   & 63.5                     & \multicolumn{1}{r|}{65.1}   & 78.4                     & 79.5                       \\

\textbf{\texttt{MUSE}}$_\mu$ &   \bf 72.2 & 	71.8 & 	 \bf 82.3 & 	 \bf 82.5 & 	 \bf 70.5 & 	 \bf 70.1 & 	 \bf 61.2 & 	62.7 & 	 \bf 71.9 & 	 \bf 71.4 & 	 \bf 64.1 & 	 \bf 65.3 & 	 \bf 78.8 & 	 \bf 80.5 \\

\end{tabular}
}
\caption{Cross-lingual word similarity results. Pearson ($r$) and Spearman ($\rho$) correlation.}
\label{tab:cross-sim}
\end{table*}

\paragraph{Analysis}

\begin{figure}[!t] 
    \includegraphics[width=\columnwidth]{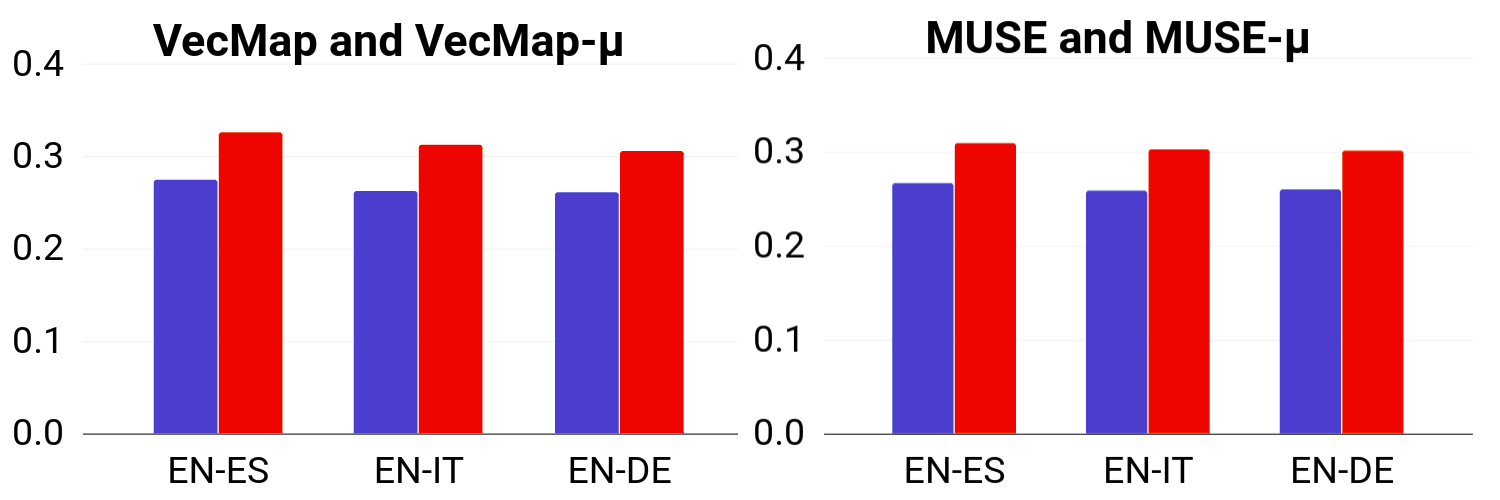} 
  \caption{Comparative average similarity between \texttt{VecMap} and \texttt{MUSE} (blue) and our proposed model (red) on the SemEval cross-lingual similarity datasets.}
  \label{fig:avsim} 
\end{figure}

In order to further understand the movements of the space with respect to the original \texttt{VecMap} and \texttt{MUSE} spaces, Figure \ref{fig:avsim} displays the average similarity values on the SemEval cross-lingual datasets (the largest among all benchmarks) of each model. As expected, the figure clearly shows how our model consistently brings the words from both languages closer on all language pairs. Furthermore, this movement is performed smoothly across all pairs, i.e., our model does not make large changes to specific words but rather small changes overall. This can be verified by inspecting the standard deviation of the difference in similarity after applying our transformation. These standard deviation scores range from 0.031 (English-Spanish for \texttt{VecMap}) to 0.039 (English-Italian for \texttt{MUSE}), which are relatively small given that the cosine similarity scale ranges from -1 to 1.

As a complement of this analysis we show some qualitative results which give us further insights on the transformations of the vector space after our average approximation. In particular, we analyze the reasons behind the higher quality displayed by our bilingual embeddings in monolingual settings. 
While \texttt{VecMap} and \texttt{MUSE} do not transform the initial monolingual spaces, our model transforms both spaces simultaneously.  In this analysis we focus on the source language of our experiments (i.e., English). We found interesting patterns which are learned by our model and help understand these monolingual gains.  
For example, a recurring pattern is that words in English which are translated to the same word, or to semantically close words, in the target language end up closer together after our transformation.  For example, in the case of English-Spanish the following pairs were among the pairs whose similarity increased the most by applying our transformation: \textit{cellphone-telephone}, \textit{movie-film}, \textit{book-manuscript} or \textit{rhythm-cadence}, which are either translated to the same word in Spanish (i.e., \textit{tel\'efono} and \textit{pel\'icula} in the first two cases) or are already very close in the Spanish space. More generally, we found that word pairs which move together the most tend to be semantically very similar and belong to the same domain, e.g., \textit{car-bicycle}, \textit{opera-cinema}, or \textit{snow-ice}.

\subsubsection{Cross-lingual hypernym discovery}
\label{hypernym_discovery}

Modeling hypernymy is a crucial task in NLP, with direct applications in diverse areas such as semantic search \cite{hoffart2014stics,roller-erk:2016:EMNLP2016}, question answering \cite{prager2008question,yahya2013robust} or textual entailment \cite{geffet2005distributional}. Hypernyms, in addition, are the backbone of lexical ontologies \cite{Yuetal2015}, which are in turn useful for organizing, navigating and retrieving online content \cite{bordea2016semeval}. Thus, we propose to evaluate the contribution of cross-lingual embeddings towards the task of hypernym discovery, i.e., given an input word (e.g., \textit{cat}), retrieve or discover its most likely (set of) valid hypernyms (e.g., \textit{animal}, \textit{mammal}, \textit{feline}, and so on). Intuitively, by leveraging a bilingual vector space condensing the semantics of two languages, one of them being English, the need for large amounts of training data in the target language may be reduced. 

\paragraph{Experimental setting}

We follow \newcite{EspinosaEMNLP2016} and learn a (cross-lingual) linear transformation matrix between the hyponym and hypernym spaces, which is afterwards used to predict the most likely (set of) hypernyms, given an unseen hyponym. Training and evaluation data come from the SemEval 2018 Shared Task on Hypernym Discovery \cite{semeval2018task9}. Note that current state-of-the-art systems aimed at modeling hypernymy \cite{shwartzetal2016,berniercolborne-barriere:2018:SemEval-2018} combine large amounts of annotated data along with language-specific rules and cue phrases such as Hearst Patterns \cite{Hearst:92}, both of which are generally scarcely (if at all) available for languages other than English. 
Therefore, we report experiments with training data only from English (11,779 hyponym-hypernym pairs), and ``enriched'' models informed with relatively few training pairs (500, 1k and 2k) from the target languages. Evaluation is conducted with the same metrics as in the original SemEval task, i.e.,  Mean Reciprocal Rank (MRR), Mean Average Precision (MAP) and Precision at 5 (P@5). These measures explain a model's behavior from complementary prisms, namely how often at least one valid hypernym was highly ranked (MRR), and in cases where there is more than one correct hypernym, to what extent they were all correctly retrieved (MAP and P@5). Finally, as in the previous experiments, we report comparative results between our proposed models and the two competing baselines (\texttt{VecMap} and \texttt{MUSE}). As an additional informative baseline, we include the highest scoring unsupervised system at the SemEval task for both Spanish and Italian (\texttt{BestUns}), which is based on the distributional models described in \newcite{shwartz2017hypernymy}.

\paragraph{Results} 

The results listed in Table \ref{tab:hyp} indicate several trends.\footnote{Note that this task is harder than hypernymy detection \cite{upadhyay2018robust}. Hypernymy detection is framed as a binary classification task, while in hypernym discovery hypernyms have to be retrieved from the whole vocabulary.} First and foremost, in terms of model-wise comparisons, we observe that our proposed alterations of both \texttt{VecMap} and \texttt{MUSE} improve their quality in a consistent manner, across most metrics and data configurations.  
In Italian our proposed model shows an improvement across all configurations. However, in Spanish \texttt{VecMap} emerges as a highly competitive baseline, with our model only showing an improved performance when training data in this language abounds (in this specific case there is an increase from 17.2 to 19.5 points in the MRR metric). This suggests that the fact that the monolingual spaces are closer in our model is clearly beneficial when hybrid training data is given as input, opening up avenues for future work on weakly-supervised learning. 
Concerning the other baseline, \texttt{MUSE}, the contribution of our proposed model is consistent for both languages, again becoming more apparent in the Italian split and in a fully cross-lingual setting, where the improvement in MRR is almost 3 points (from 10.6 to 13.3). Finally, it is noteworthy that even in the setting where no training data from the target language is leveraged, all the systems based on cross-lingual embeddings outperform the best unsupervised baseline, which is a very encouraging result with regards to solving tasks for languages on which training data is not easily accessible or not directly available. 

\begin{table}[!t]
\centering
\renewcommand{\arraystretch}{1.15}
\resizebox{\columnwidth}{!}{
\begin{tabular}{c|l|rrr|rrr}
\multicolumn{1}{c|}{\multirow{2}{*}{\begin{tabular}[c]{@{}c@{}}Train\\ data\end{tabular}}} & \multicolumn{1}{c|}{\multirow{2}{*}{Model}} & \multicolumn{3}{c|}{\texttt{\textbf{Spanish}}}                                                   & \multicolumn{3}{c}{\texttt{\textbf{Italian}}}                                                  \\ \cline{3-8} 
\multicolumn{1}{c|}{}                                                                      & \multicolumn{1}{l|}{}                       & \multicolumn{1}{c}{{\small{MAP}}} & \multicolumn{1}{c}{{\small{MRR}}} & \multicolumn{1}{c|}{{\small{P@5}}} & \multicolumn{1}{c}{{\small{MAP}}} & \multicolumn{1}{c}{{\small{MRR}}} & \multicolumn{1}{c}{{\small{P@5}}} \\ \hline
\multicolumn{1}{c|}{-}                                                                     & \textbf{\texttt{BestUns}}                                     & 2.4                      & 5.5                      & 2.5                      & 3.9                      & 8.7                      & 3.9                     \\ \hline
\multirow{4}{*}{\textbf{EN}}                                                                        & \textbf{\texttt{VecMap}}                                          &  \bf 6.4                      &  \bf 16.5                     &  \bf 6.0                      & 4.5                      & 10.6                     & 4.3                     \\
                                                                                           & \textbf{\texttt{VecMap}$_\mu$}                                    & 6.1                      & 15.4                     & 5.7                      & \bf  5.6                      &  \bf 13.3                     &  \bf 5.4              \\ \cline{2-8} 
                                                                                           & \textbf{\texttt{MUSE}}                                        & 5.9                      & 14.1                     & 5.5                      & 4.9                      & 11.1                     & 4.7                     \\
                                                                                           & \textbf{\texttt{MUSE}$_\mu$}                                 &  \bf 6.2                      &  \bf 14.8                     &  \bf 5.8                      &  \bf 5.1                      &  \bf 11.7                     &  \bf 4.9              \\ \hline
\multirow{4}{*}{\begin{tabular}[c]{@{}c@{}}\textbf{EN}\\ \textbf{+}\\ \textbf{500}\end{tabular}}                      & \textbf{\texttt{VecMap}}                                          &  \bf 7.3                      &  \bf 18.2                     &  \bf 7.0                      & 6.1                      & 14.0                     & 5.8                     \\
                                                                                           & \textbf{\texttt{VecMap}$_\mu$}                                 & 7.0                      & 17.6                     & 6.6                      &  \bf 6.8                      &  \bf 16.2                     &  \bf 6.4           \\ \cline{2-8} 
                                                                                           & \textbf{\texttt{MUSE}}                                        & 6.4                      & 15.9                     & 6.1                      & 5.3                      & 12.0                     & 5.0                     \\
                                                                                           & \textbf{\texttt{MUSE}$_\mu$}                             &  \bf 6.9                      &  \bf 16.9                     &  \bf 6.6                      &  \bf 6.0                      &  \bf 13.4                     &  \bf 5.7              \\ \hline
\multirow{4}{*}{\begin{tabular}[c]{@{}c@{}}\textbf{EN}\\ \textbf{+}\\ \textbf{1k}\end{tabular}}                       & \textbf{\texttt{VecMap}}                                          &  \bf 7.9                      &  19.2                     &  \bf 7.6                      & 7.0                      & 16.4                     & 6.6                     \\
                                                                                           & \textbf{\texttt{VecMap}$_\mu$}                                    & 7.8                      & 19.2                     & 7.4             &  \bf 7.5                      &  \bf 18.1                     &  \bf 7.0          \\ \cline{2-8} 
                                                                                           & \textbf{\texttt{MUSE}}                                        & 7.2                      & 17.3                     & 6.9                      & 6.2                      & 13.8                     & 5.8                     \\
                                                                                           & \textbf{\texttt{MUSE}$_\mu$}                                 &  \bf 7.8                      &  \bf 18.8                     &  \bf 7.5                      &  \bf 6.5                      &  \bf 14.2                     &  \bf 6.3                   \\ \hline
\multirow{4}{*}{\begin{tabular}[c]{@{}c@{}}\textbf{EN}\\ \textbf{+}\\ \textbf{2k}\end{tabular}}                       & \textbf{\texttt{VecMap}}                                          & 8.0                      & 19.1                     & 7.7                      & 8.2                      & 19.1                     & 7.5                     \\
                                                                                           & \textbf{\texttt{VecMap}$_\mu$}                                &  \bf 8.2                      &  \bf 19.9                     &  \bf 7.9                      &  \bf 8.7                      &  \bf 20.7                     &  \bf 8.1         \\ \cline{2-8} 
                                                                                           & \textbf{\texttt{MUSE}}                                        & 7.2                      & 17.2                     & 6.8                      & 7.2                      & 15.8                     & 7.0                     \\
                                                                                           & \textbf{\texttt{MUSE}$_\mu$}                              &  \bf 8.3                      &  \bf 19.5                     &  \bf 8.0                      &  \bf 7.6                      &  \bf 17.0                     &  \bf 7.2                
\end{tabular}}
\caption{Results on the hypernym discovery task.}
\label{tab:hyp}
\end{table}
 
\paragraph{Analysis}

A manual exploration of the results obtained in cross-lingual hypernym discovery reveals a systematic pattern when comparing, for example, \texttt{VecMap}  and our model. It was shown in Table \ref{tab:hyp} that the performance of our model gradually increased alongside the size of the training data in the target language until surpassing \texttt{VecMap} in the most informed configuration (i.e., EN+2k). Specifically, our model seems to show a higher presence of generic words in the output hypernyms, which may be explained by these being closer in the space. In fact, out of 1000 candidate hyponyms, our model correctly finds \textit{person} 143 times, as compared to the 111 of \texttt{VecMap}, and this systematically occurs with generic types such as \textit{citizen} or \textit{transport}.
Let us mention, however, that the considered baselines perform remarkably well in some cases. For example, the English-only \texttt{VecMap} configuration (EN), unlike ours, correctly discovered the following hypernyms for \textit{Francesc Maci\`{a}} (a Spanish politician and soldier): \textit{politician}, \textit{ruler}, \textit{leader} and \textit{person}. These were missing from the prediction of our model in all configurations until the most informed one (EN+2k).

\section{Conclusions and Future Work}

We have shown how to refine bilingual word embeddings by applying a simple transformation which moves cross-lingual synonyms closer towards their average representation. 
Before applying this strategy, we start by aligning the monolingual embeddings of the two languages of interest.
For this initial alignment, we have considered two state-of-the-art methods from the literature, namely \texttt{VecMap} \cite{artetxe-labaka-agirre:2017:Long} and \texttt{MUSE} \cite{conneau2018word}, which also served as our baselines.
Our approach is motivated by the fact that these alignment methods do not change the structure of the individual monolingual spaces. However, the internal structure of embeddings is, at least to some extent, language-specific, and is moreover affected by biases of the corpus from which they are trained, meaning that after the initial alignment significant gaps remain between the representations of cross-lingual synonyms.
We tested our approach on a wide array of datasets from different tasks (i.e., bilingual dictionary induction, word similarity and cross-lingual hypernym discovery) with state-of-the-art results. 

This paper opens up several promising avenues for future work.
First, even though both languages are currently being treated symmetrically, the initial monolingual embedding of one of the languages may be more reliable than that of the other. In such cases, it may be of interest to replace the vectors $\vec{\mu}_ {w,w'}$ by a weighted average of the monolingual word vectors. 
Second, while we have only considered bilingual scenarios in this paper, our approach can naturally be applied to scenarios involving more languages.
In this case, we would first choose a single target language, and obtain alignments between all the other languages and this target language. 
To apply our model, we can then simply learn mappings to predict averaged word vectors across all languages.
Finally, it would also be interesting to use the obtained embeddings in downstream applications such as language identification or cross-lingual sentiment analysis, and extend our analysis to other languages, with a particular focus on morphologically-rich languages (after seeing our success with Finnish), for which the bilingual induction task has proved more challenging for standard cross-lingual embedding models~\cite{sogaard2018limitations}. 

\section*{Acknowledgments}

Yerai Doval is funded by the Spanish Ministry of Economy, Industry and Competitiveness (MINECO) through project FFI2014-51978-C2-2-R, and by the Spanish State Secretariat for Research, Development and Innovation (which belongs to MINECO) and the European Social Fund (ESF) under a FPI fellowship (BES-2015-073768) associated to project FFI2014-51978-C2-1-R. Jose Camacho-Collados, Luis Espinosa-Anke and Steven Schockaert have been supported by ERC Starting Grant 637277.

\bibliographystyle{acl_natbib_nourl}

\end{document}